\documentclass[runningheads]{llncs}

\usepackage{eccv}

\usepackage{eccvabbrv}

\usepackage{graphicx}
\usepackage{booktabs}
\usepackage{comment}

\usepackage[accsupp]{axessibility}  

\usepackage{hyperref}

\usepackage{orcidlink}

\usepackage{bm}
\usepackage{siunitx}
\usepackage{threeparttable}
\usepackage{booktabs} 
\usepackage{wrapfig} 
\usepackage{multibib}
\newcites{supp}{Supplementary References}

\usepackage{xspace}

\DeclareMathOperator*{\argmin}{arg\,min}
\DeclareMathOperator*{\argmax}{arg\,max}

\DeclareSIUnit[prefixes-as-symbols=false]{\rpm}{rpm}
\usepackage{tablefootnote}

\usepackage[acronym]{glossaries}

\newcommand{\newgcsacronym}[3]{%
  \newacronym{#1}{#2}{#3}%
  \expandafter\newcommand\csname #1\endcsname{\gls{#1}\xspace}%
}
\usepackage{bbding} 
\newgcsacronym{gcs}{GCS}{gaze control system}
\newgcsacronym{evs}{EVS}{event-based vision sensor}
\newgcsacronym{aps}{APS}{active pixel sensor}
\newgcsacronym{scmax}{s-CMax}{spherical contrast maximization}
\newgcsacronym{iwe}{IWE}{image of warped events}
\newgcsacronym{cnn}{CNN}{convolutional neural network}
\newgcsacronym{bpms}{BPMS}{ball position measurement system}
\newgcsacronym{fov}{FoV}{field of view}

\newif\ifshowfigures
\showfigurestrue

\newif\ifshowtables
\showtablestrue

\makeatletter
\renewcommand\section{\@startsection{section}{1}{\z@}%
  {-12pt \@plus -2pt \@minus -2pt}
  {6pt \@plus 2pt}
  {\normalfont\large\bfseries}}
\renewcommand\subsection{\@startsection{subsection}{2}{\z@}%
  {-8pt \@plus -2pt \@minus -1pt}%
  {4pt \@plus 1pt}%
  {\normalfont\normalsize\bfseries}}
\renewcommand\subsubsection{\@startsection{subsubsection}{3}{\z@}%
  {-6pt \@plus -1pt \@minus -1pt}%
  {3pt \@plus 1pt}%
  {\normalfont\normalsize\bfseries}}
\makeatother

\begin{document}

\title{Event-based Gaze Control System \\
for Accurate Real-time Spin Estimation \\ 
in Professional Ball Games}

\titlerunning{Event-based Spin Estimation}

\author{Yunpu Hu\inst{1}\textsuperscript{*}\orcidlink{0000-0003-3623-4522} \and
        Fabian Schilling\inst{2}\textsuperscript{*}\orcidlink{0000-0003-3787-5137}\and
        Valentina Cavinato\inst{3}\textsuperscript{*}\orcidlink{0009-0007-4500-5592}\and\\
        Asude Aydin\inst{2}\orcidlink{0000-0003-2059-3387}\and
        Agis Politis\inst{2}\and
        Ricardo Tapiador Morales\inst{2}\orcidlink{0000-0002-7268-2915}\and\\
        Kirk Y.W. Scheper\inst{3}\orcidlink{0000-0003-2770-5556}\and
        Peter D\"urr\inst{2}\orcidlink{0000-0002-3840-5009}\and
        Naoya Takahashi\inst{2}\Envelope\orcidlink{0000-0001-8553-4797}%
}

\authorrunning{Y.~Hu et al.}

\institute{Sony AI, Tokyo, Japan \and
        Sony AI, Z\"urich, Switzerland \and
       Sony Advanced Visual Sensing, Sony Europe Ltd., Zurich, Switzerland.}
\maketitle
{\renewcommand\thefootnote{}\footnotetext{$^*$These authors contributed equally.}}

\begin{abstract}
Spin plays a crucial role in many ball sports due to its effect on the trajectory of the ball.
Vision-based estimation of the ball's spin during a game with conventional cameras is challenging due to the ball's small size, high speed, and fast rotation.
To address these challenges, we propose an event-based active vision system that can track unmodified balls and measure their spin in real time.
The system consists of an event camera for its high temporal resolution and minimal motion blur, high-speed pan/tilt galvanometer mirrors to keep the ball in the field of view, and a low-latency focus-tunable telephoto lens to increase the spatial resolution on the ball and keep it in focus.
To track the ball, we use a hybrid approach that combines 2D event-based detection for centering and 3D positions from a ball localization system for re-initialization. 
For high-accuracy spin estimation, we propose an offline method that performs contrast maximization on the sphere (s-CMax).
This method achieves state-of-the-art accuracy on static balls across multiple sports (table tennis, baseball, tennis, and golf), with mean magnitude and axis errors of 1.2\% and 1.5 degrees, respectively.
We then develop a low-latency online method for table tennis as a case study in real-time applications.
This method uses an uncertainty-aware convolutional neural network trained on pseudo-ground-truth spin labels from the offline approach, combined with a GPU-accelerated batch implementation of contrast maximization for refinement.
We demonstrate reliable tracking and spin estimation with a three-view setup during professional table tennis matches, with high accuracy (8.8\% magnitude and 6.4 degrees axis mismatch w.r.t. the offline method), 3 ms latency, and 750 Hz throughput.

\keywords{Event-based cameras \and Object pose estimation and tracking \and Computer vision for robotics}
\end{abstract}

    
\section{Introduction}
\label{sec:intro}

Spin plays a key role in many ball sports due to its effect on the ball's flight and contact dynamics.
Athletes routinely apply spin to control the ball's flight path, deceive opponents, or gain a tactical advantage.
Accurate spin estimation is thus of vital importance for performance analysis, sports statistics, physics modeling \cite{Conti2026}, trajectory prediction \cite{Bi2026}, and competitive robotic gameplay~\cite{durr_outplaying_2026}.
For example, in table tennis, rallies can feature balls spinning at magnitudes of up to \SI{9}{\kilo\rpm} \cite{tang_speed_2002}.
In baseball, curve balls can reach spin rates up to \SI{2.5}{\kilo\rpm} \cite{hashimoto_relationship_2023}, tennis serves can impart spins up to \SI{5}{\kilo\rpm} \cite{kashiwagi_differences_2021}, and golf wedge shots can exceed \SI{10}{\kilo\rpm} \cite{usga_second_2007}.

\ifshowfigures
\begin{figure}[!t]
    \centering
    \includegraphics[width=0.9\linewidth]{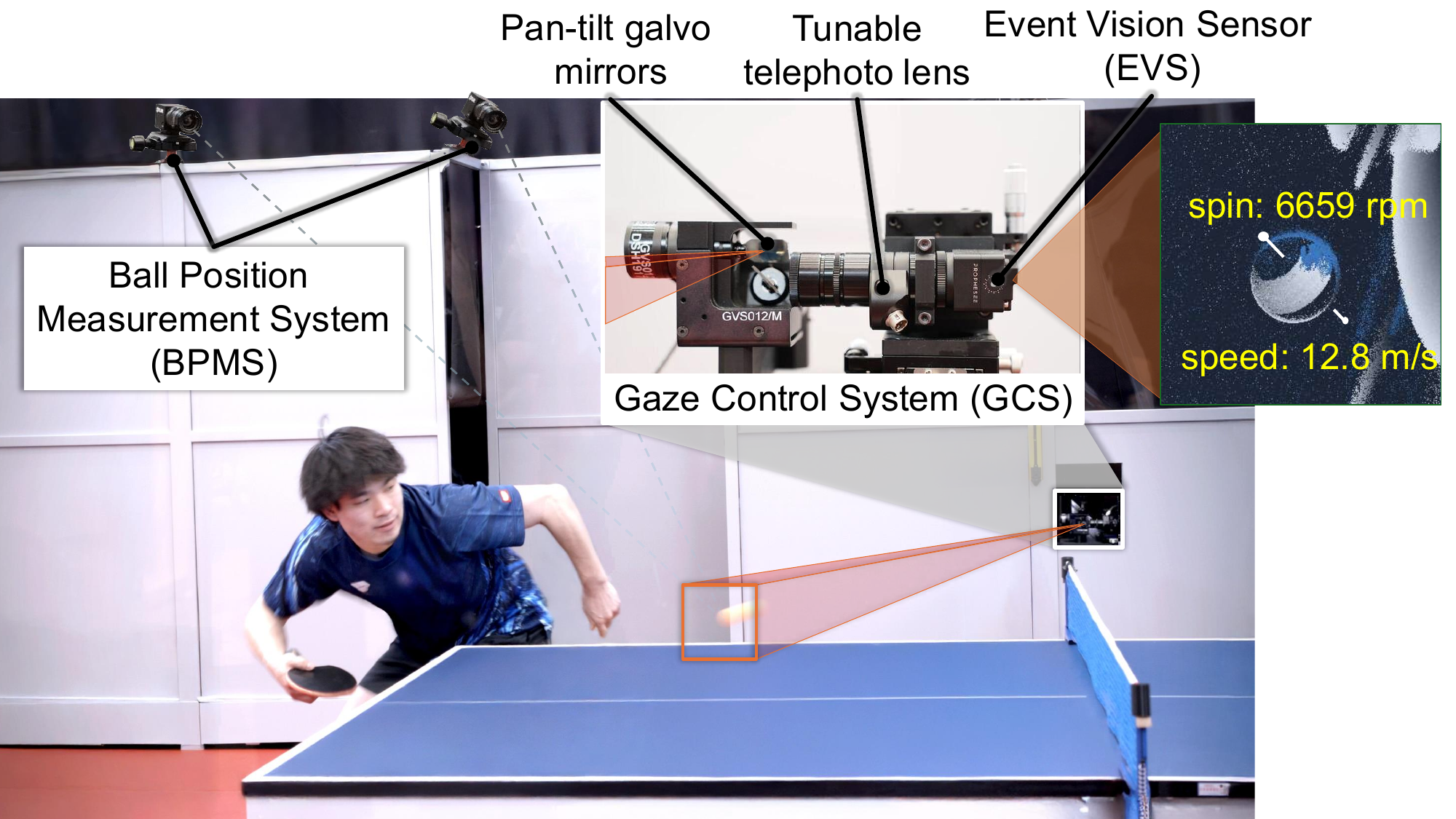}
    \caption{The Gaze Control System (GCS) actively tracks the ball and measures its spin.}
    \label{fig:main_figure}
\end{figure}
\fi

Vision-based ball spin estimation is challenging mainly for two reasons.
First, balls are typically small, spanning only a few pixels in the \fov of a wide-angle camera and making surface features hard to observe; a narrow-angle camera, in contrast, suffers from shallow depth-of-field and limited coverage.
Second, balls' motion combines high linear velocities with high spins.
For a frame-based camera, the high image-plane velocity can introduce motion blur even at short exposures, and the spin often exceeds the Nyquist limit set by the frame rate, causing magnitude underestimation through temporal aliasing.
While an \evs~\cite{gallego_event-based_2022} can alleviate blur and aliasing, the high linear velocity typically pollutes the events and obscures the spin.

We address the above challenges with the \gcs, an event-based active vision system that tracks unmodified balls and measures their spin with high accuracy in real time.
We employ an \evs for its high temporal resolution, which avoids aliasing and reduces motion blur.
We equip the \evs with a telephoto lens for higher spatial resolution on the ball surface.
The focal power of the lens is electrically tuned with \SI{3}{\milli\second} response time to keep the ball in sharp focus despite the shallow depth of field resulting from the long focal length.
The system utilizes two precisely controlled pan/tilt galvanometer mirrors that actively control the camera's gaze direction, to keep the fast-moving ball in its \fov, and to ensure that the generated events primarily encode the ball spin, by compensating for the ball's translational motion with the mirror motion.

Despite its advantages, operating the \gcs in a highly dynamic setting poses unique challenges.
The ball's high speed and the telephoto lens's narrow \fov make robust tracking difficult, especially when the ball abruptly changes direction upon contact (e.g., with a racket, table, bat, or club).
We address this with a hybrid tracking strategy: a \bpms controls the mirrors and tunable lens, while the 2D ball position from the active \evs view provides secondary feedback, compensating for latency and noise in the \bpms, keeping the ball centered, and preventing it from exiting the \fov.

To estimate the spin from these events, we build on contrast maximization~\cite{gallego_unifying_2018,gallego_focus_2019}, a model-based framework that recovers motion parameters by maximizing the sharpness of an \iwe~\cite{gallego_accurate_2017}.
While recent works~\cite{nakabayashi_event-based_2024,sato_time-consistent_2024,gossard_table_2024} have explored event-based spin estimation, they employ static, wide-angle cameras, which limits spatial resolution on the ball and conflates translational and rotational motion in the captured events.
Since the \gcs compensates for the ball's translational motion via active mirror tracking, the events captured by the \evs primarily encode the ball's rotation, which addresses the conflation problem.
Nevertheless, prior methods apply contrast maximization on a planar image projection, which cannot faithfully represent the geometry of a spinning sphere: events generated across the curved ball surface get projected onto a flat plane, causing depth ambiguities and concentrating contrast near the ball's silhouette rather than distributing it uniformly over the surface.
We address this with a spherical formulation of contrast maximization that models events directly on the ball's surface, and develop two complementary estimation modes: an offline method for high-accuracy spin extraction, and a real-time pipeline that combines an uncertainty-aware \cnn with GPU-accelerated contrast maximization for refinement.

Our main contributions can be summarized as follows:
\begin{itemize}
    \item An event-based active vision system with a focus-tunable telephoto lens and actuated pan/tilt mirrors, capable of smoothly tracking a ball at a mirror angular velocity of \SI{142.5}{\radian\per\second} while focusing on its surface during a game.
    \item A high-accuracy offline spin estimation method based on \scmax that generalizes across multiple sports (table tennis, baseball, tennis, and golf), with mean magnitude and axis errors of \SI{1.2}{\percent} and \SI{1.5}{\degree} on spinner-mounted balls, outperforming prior event-based methods~\cite{nakabayashi_event-based_2024,gossard_table_2024}. We further validate the method on in-flight balls actively tracked by the \gcs, achieving \SI{2.1}{\percent} magnitude and \SI{5.4}{\degree} axis error.
    \item A low-latency online spin estimation method featuring an uncertainty-aware convolutional neural network trained on pseudo-ground-truth spin labels from the offline approach, combined with a GPU-accelerated batch implementation of \scmax for refinement.
    \item A method for simultaneous tracking and spin estimation of unmodified balls from multiple actively controlled views, demonstrating reliable tracking and spin estimation during professional table tennis games with \SI{750}{\hertz} throughput, \SI{3}{\milli\second} latency, and \SI{8.8}{\percent} magnitude and \SI{6.4}{\degree} axis mismatch w.r.t.\ the offline method.
\end{itemize}

\section{Related work}
\label{sec:related_work}

\textit{Active gaze control for high-speed tracking.}
Capturing fast-moving objects with high spatial resolution requires active vision systems that steer the camera's gaze in real time.
The Saccade Mirror line of work~\cite{okumura_high-speed_2011,okumura_1_2015,iida_saccade_2016,miyashita_saccade_2025} introduced galvanometer-based optical gaze controllers that achieve high-speed pan/tilt tracking without physically moving the camera.
Sueishi~\etal~\cite{sueishi_high-speed_2025} coupled a similar galvanometer system with multi-exposure imaging for spin measurement of dotted table tennis balls.
Notably, all of these systems rely on frame-based cameras, and only~\cite{sueishi_high-speed_2025} addresses spin estimation, albeit requiring modified balls with custom dot patterns.
Our work extends this line of research by pairing galvanometer mirrors with an event camera and a focus-tunable telephoto lens, enabling simultaneous active tracking and spin estimation of unmodified balls.

\textit{Ball spin estimation with frame-based cameras.}
Spin estimation methods can be broadly grouped by how they observe the ball.
Trajectory-based approaches~\cite{chen_dynamic_2010,huang_trajectory_2011,su_trajectory_2013,tebbe_spin_2020,kumar2025ball} infer spin indirectly from the Magnus effect on the ball's trajectory, but require highly accurate position measurements, depend on calibrated aerodynamic models (Magnus and drag coefficients vary with velocity and air conditions), and offer limited observability of the full spin axis.
Image-based methods instead observe the ball surface directly.
Some rely on custom markers~\cite{tamaki_measuring_2004,theobalt_pitching_2004,furuno_study_2009,tamaki_estimating_2012,gossard_spindoe_2023,sueishi_high-speed_2025}, limiting practical applicability.
Methods for unmodified balls typically track surface texture across frames~\cite{glover_tracking_2014,zhang_spin_2014,zhang_real-time_2015,tamaki_spin_2024,tebbe_spin_2020}, or recover 3D spin from motion blur in a single frame~\cite{boracchi_estimation_2009}, though the latter requires carefully controlled exposure.
Closest to our approach, Zhang~\etal~\cite{zhang_spin_2014,zhang_real-time_2015} paired a wide-angle multi-camera system for 3D localization with pan/tilt telephoto cameras for spin observation, estimating spin by segmenting the logo across successive frames.
However, their evaluation reaches only \SI{2000}{\rpm} and relies on indirect, trajectory-based error metrics.
More broadly, all frame-based methods face fundamental limits at high spin: even short exposures introduce motion blur, and the discrete frame rate imposes a Nyquist limit beyond which temporal aliasing underestimates the spin.

\textit{Event-based spin estimation.}
Event cameras~\cite{gallego_event-based_2022} respond asynchronously to per-pixel brightness changes with microsecond resolution, producing substantially reduced motion blur.
These properties are well suited for observing fast-spinning balls.
Nakabayashi~\etal~\cite{nakabayashi_event-based_2024,sato_time-consistent_2024} were the first to apply contrast maximization (CMax)~\cite{gallego_unifying_2018,gallego_focus_2019} to ball spin estimation.
Their work demonstrated results on simulated data, although with manually provided ball detections and without reporting runtime or real-world quantitative evaluation.
Gossard~\etal~\cite{gossard_table_2024} proposed an optic-flow-based approach for event-based spin estimation that leverages the periodicity in the event stream to set velocity constraints.
However, all existing event-based methods employ static, wide-angle cameras, which severely limits the spatial resolution on the ball and conflates translational motion with spin, a fundamental challenge for in-flight estimation.
Our work addresses both limitations through active tracking with a telephoto lens, proposes a spherical formulation of CMax for more accurate rotation estimation, and develops a real-time pipeline combining an uncertainty-aware \cnn with GPU-accelerated CMax refinement.

\section{Methods}
\label{sec:methods}

\subsection{Overview}
\label{sec:method_overview}

The present system is designed to actively track a ball and estimate its spin, supporting both an offline mode for high-accuracy estimates and a real-time mode for time-sensitive applications (Fig.~\ref{fig:method_overview}).
Both modes share the same active tracking pipeline.
We estimate the state of the ball using 3D ball positions obtained continuously from a \bpms (Sec.~\ref{sec:bpms}) to control the \gcs (Sec.~\ref{sec:gcs}).
We detect the ball asynchronously within the \evs event stream to track it and to filter events for spin estimation (Sec.~\ref{sec:ball_detection}).
The hybrid tracking framework uses the 3D ball positions and 2D ball detections to provide smooth and centered tracking (Sec.~\ref{sec:ball_tracking}).
Using this active tracking framework, we run \scmax offline to extract high-accuracy pseudo-ground-truth spin labels (Sec.~\ref{sec:method_offline_spin_estimation}), on which we train a low-latency uncertainty-aware \cnn for real-time operation (Sec.~\ref{sec:method_online_spin_estimation}); the resulting estimates are fed back to improve the tracking (Sec.~\ref{sec:state_estimation}).

\ifshowfigures
\begin{figure}[!t]
    \centering
    \includegraphics[width=1\linewidth]{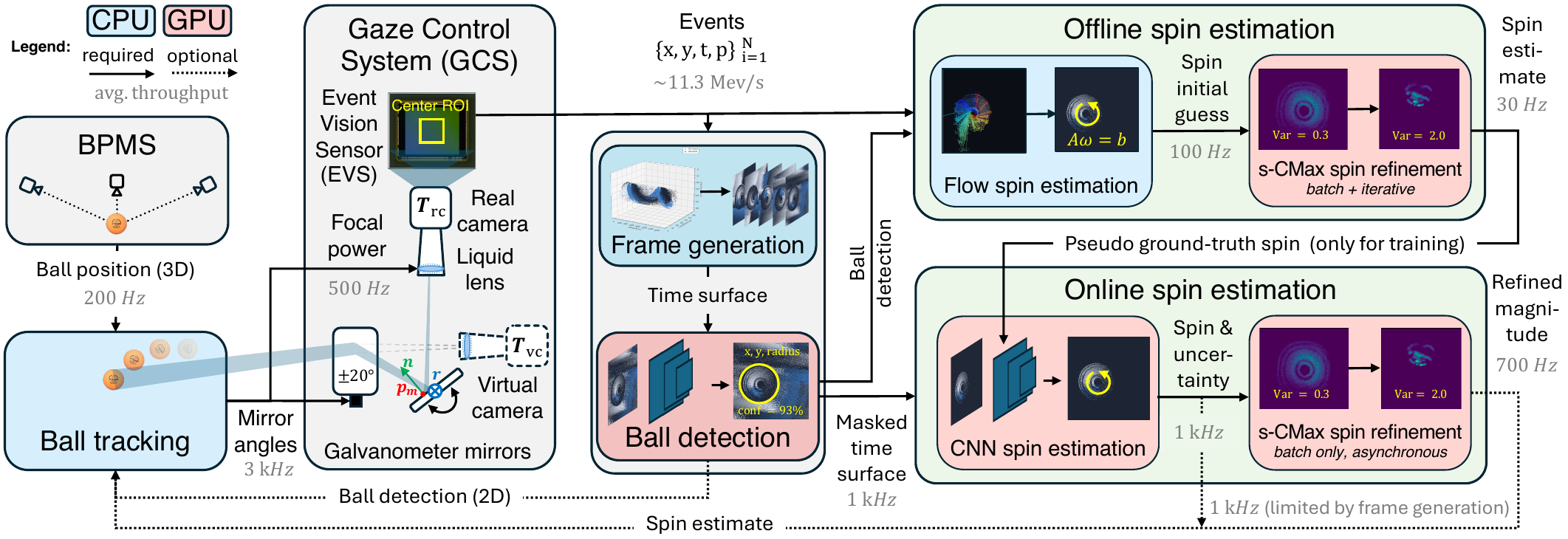}
    \caption{Overview of the information flow in the \gcs.}
    \label{fig:method_overview}
    \vspace{-15pt}

\end{figure}
\fi

\subsubsection{Ball position measurement system (3D)}
\label{sec:bpms}

The \bpms provides a continuous stream of 3D ball position measurements with appropriate coverage, accuracy, latency, and update rate for the active tracking system.
In theory, any system providing 3D position measurements can be used, e.g., a multi-camera system, depth sensors, or radar-based systems.
In this work, the real table tennis game recordings used a nine-camera \aps \bpms~\cite{durr_outplaying_2026}, where each camera detects the 2D ball position and the 3D position is estimated through triangulation.
The spinner dataset (Sec.~\ref{sec:exp_spin_estimation_ball_spinner}) was recorded using a monocular camera tracking an AprilTag.

\subsubsection{Gaze control system (GCS)}
\label{sec:gcs}

The \gcs comprises three main hardware components: 1) an \evs, 2) a focus-tunable telephoto lens, and 3) pan/tilt galvanometer mirrors (Fig.~\ref{fig:main_figure}).
We choose an event camera for its high temporal resolution and minimal motion blur, which are critical for resolving the fine surface texture of a fast-spinning ball without aliasing.
The event camera is a Prophesee EVK4 featuring a Sony/Prophesee IMX 636 sensor \cite{finateu_1280720_2020}.
We use a centered hardware ROI of $320 \times 320$ to avoid event loss caused by bandwidth limitation in high event rate scenes.
The telephoto lens is used to increase the spatial resolution on the ball; however, its narrow field of view and shallow depth of field demand active focus control as the ball distance changes rapidly.
We use an Optotune EL-10-30 focus-tunable liquid lens because it offers a fast step response on the order of milliseconds.
The lens is embedded in a custom optical design with \SI{50}{\milli\meter} focal length and an effective field of view of $\sim1.8^{\circ}$.
To steer the camera's gaze, we use galvanometer mirrors rather than physically rotating the camera.
The galvanometer is a Thorlabs GVS012/M which provides a maximal mechanical scan angle of $\pm 20^{\circ}$ for each pan and tilt mirror.
Since each mirror rotates only a small, lightweight optical element, its rotational inertia is extremely low, enabling a small-angle step response of \SI{400}{\micro\second} and a sustained mirror speed of \SI{142.5}{\radian\per\second} (roughly \SI{855}{\meter\per\second} target lateral motion at \SI{3}{\meter} distance), well exceeding the speeds in professional ball sports.
The system is calibrated in four stages, all referenced to a common frame via AprilTags~\cite{wang_apriltag_2016}: camera intrinsics (\aps and reconstruction-based \evs~\cite{muglikar_how_2021}), mirror extrinsics~\cite{sturm_how_2006}, lens distance-to-focal-power map, and \bpms-to-\gcs transform (Suppl.~Sec.~A.3).
Spin estimation runs in the virtual camera frame and is largely insensitive to calibration error: \scmax and the \cnn operate in 2D image space, so intrinsics do not propagate, while mirror and extrinsic errors are bounded by the narrow \fov ($\pm0.9^{\circ}$, well below the prior state-of-the-art axis error of \SI{5.0}{\degree}(Tab.~\ref{tab:method_comparison_new_details}))---in practice we observe $\sim$\SI{1}{px} reprojection error ($5.6\times10^{-3}\,^{\circ}$).
Any residual error is shared across methods and thus cancels in our comparisons.

\textit{Multi-system fusion.}
In practice, the spin-relevant pattern may be visible from only from a limited subset of viewpoints (e.g., the table tennis ball; Fig.~\ref{fig:four_balls}), so it can be necessary to deploy multiple \gcs placed to maximize the visible ball-surface area.
In our table-tennis game setup~\cite{durr_outplaying_2026}, we fuse per-view estimates by selecting the prediction with the lowest uncertainty, improving robustness to occlusions and expanding the observable workspace (Suppl.~Sec.~A.7).

\subsubsection{Event-based frame generation and ball detection (2D)}
\label{sec:ball_detection}

We detect the ball in the \evs frame
1) to isolate it from the background for spin estimation and 2) to provide a direct feedback signal for active ball tracking.
Since the appearance of the ball within the event stream can change drastically based on the mirror motion, we opt for a deep-learning-based detector.
We convert the event stream to a polarity-separated time surface representation \cite{benosman_event-based_2014} since it preserves motion information and has a small memory footprint for inference, while allowing convenient truncation during online operation.
We use a YOLO object detection model \cite{wang_yolov9_2024} to predict a single bounding circle ($x$, $y$ pixel location, radius, and confidence) from the time surface with an event accumulation time of $\SI{5}{\milli\second}$.
For details on the dataset and training methodology, please refer to the supplementary material (Suppl.~Sec.~A.5).

\subsection{State estimation and ball tracking}
\label{sec:ball_tracking}

The hybrid ball tracking method combines complementary open-loop position-based and closed-loop image-based visual servoing techniques that are based on a real-time state estimate.
The position-based method takes 3D ball positions obtained from the \bpms (Sec.~\ref{sec:bpms}) whereas the image-based method takes 2D ball detections from the \evs (Sec.~\ref{sec:ball_detection}).
The tracking method controls the mirror angles of the galvo and the focal power of the liquid lens to keep the ball within the frame and in sharp focus, respectively.

\subsubsection{State estimation}
\label{sec:state_estimation}

We employ an improved Nakashima model \cite{Conti2026,nakashima_modeling_2011} to estimate the ball states, accounting for factors such as air drag, gravity, and the Magnus effect.
The model parameters are fitted using datasets collected from real table tennis games.
For each free-flight trajectory, the filter initializes its states, including ball position and velocity, using the first few ball position inputs. 
The filter allows for smooth tracking through latency compensation by predicting future ball positions and by controlling the mirrors incrementally in the absence of a ball position input.

\textit{Contact handling in tracking.} 
Most contacts in ball games can be regarded as the start of a new free-flight trajectory and trigger a tracking re-initialization.
However, for some contacts, such as table contacts in table tennis, we predict the post-contact trajectory from the pre-contact ball state since the normal direction of the contact surface is known~\cite{nakashima_modeling_2010,nakashima_modeling_2011}.
Note that this model requires the pre-contact spin as the input, for which we are feeding the real-time spin estimates.

\textit{Uncertainty-aware spin filter.} 
To temporally smooth the spin estimates, we aggregate spin predictions from online spin estimation by maintaining a sliding history buffer of up to $45$ estimates along with their uncertainties.
Each prediction is weighted by the inverse of its uncertainty value, effectively prioritizing more confident estimates in the aggregation.
Assuming the spin is constant between contacts, the filter rejects outliers based on 1) median absolute deviation thresholding over spin, and 2) uncertainty thresholding. 
The filter state resets only upon a detected contact, ensuring stable and consistent estimates even when the logo is temporarily occluded.
The final estimate is the inverse-uncertainty-weighted mean of the inlier set.

\subsubsection{Open-loop position-based tracking} 
\label{sec:tracking_open_loop}

The \gcs receives 3D ball state estimates, which are continuously updated by the \bpms.
Both systems are synchronized using a centralized hardware trigger generator, which allows estimation of the average sensing latency from exposure time to mirror control time.
For each 3D ball position input $\mathbf{p}_t$, the mirror angles $\phi_0$ and $\phi_1$ are computed by solving the optimization problem
\begin{equation}
\label{eq:3dtracking}
    \phi_0, \phi_1 = \argmin_{\phi_0, \phi_1} \left\| z_t^{-1} K_c T_\text{vc} \mathbf{p}_t - \left[c_x, c_y, 1\right]^\top \right\|
\end{equation}
where $K_c$ is the intrinsic matrix, $z_t$ is the z-coordinate of the ball position, and $(c_x,c_y)$ are the center pixel coordinates of the event camera.
This optimization problem (Eq.~\ref{eq:3dtracking}) can be solved using gradient descent.
The lens focal power is simultaneously adjusted based on the target distance to the virtual camera.

\subsubsection{Closed-loop image-based tracking} 
\label{sec:tracking_closed_loop}

In practice, position noise and temporal instability in the \bpms may result in tracking failures, which motivates the use of 2D ball detection in the \evs frame.
We 1) reinitialize the 3D ball state estimator using new \bpms inputs if the ball is not detected in the \evs frame, and 2) adjust the 3D ball position estimate based on the detection, ensuring that the ball remains centered in the frame.
For 2), we compute 
$
   \Delta\mathbf{p}_t= z_tR_{vc}^{-1}K_c^{-1}\, \left[\Delta u, \Delta v, 1 \right]^\top
$, 
where $(\Delta u, \Delta v)$ represent the 2D coordinate differences from the image center, and $\Delta \mathbf{p}_t$ denotes the difference between the expected and detected 3D ball positions.
When computing, we assume that the depth $z_t$ has minimal difference from the filter ball state. $\Delta\mathbf{p}_t$ is then added to the filter state to lock the ball to the image center.
This hybrid tracker is explicitly designed to tolerate \bpms noise, latency, and brief dropouts.
We have verified that a simple two-camera stereo rig (\SI{30}{\centi\meter} baseline) suffices for tracking at \qtyrange{3}{20}{\meter} without noticeable degradation.

\subsection{Offline spin estimation}
\label{sec:method_offline_spin_estimation}

The offline estimator runs in two stages: optical flow first initializes spin from local event motion on the detected ball; this initialization accelerates convergence and reduces the risk that the subsequent non-convex \scmax optimization settles in an incorrect local optimum. \scmax then refines this estimate by maximizing event contrast on the sphere.

\subsubsection{Optical flow spin estimation initialization}
\label{sec:method_flow_spin_estimation}

For this initialization, we compute 2D flow vectors on masked ball events using plane fitting \cite{benosman_event-based_2014}, normalize them by the ball radius, and lift them to 3D tangential velocities $\mathbf{v}_k$ using the spherical geometry.
Since the tangential velocity at position $\mathbf{x}_k$ on the sphere relates to spin via $\mathbf{v}_k = \boldsymbol{\omega} \times \mathbf{x}_k = -[\mathbf{x}_k]_\times \boldsymbol{\omega}$ (where $[\mathbf{x}_k]_\times$ is the skew-symmetric matrix of $\mathbf{x}_k$), we construct an overdetermined system of linear equations.
Since each optical flow measurement provides only a scalar constraint along the observed flow direction (normal flow constraint \cite{zhu_event-by-event_2020}), we project each observation model using $P_k = \mathbf{v}_k\mathbf{v}_k^T / (\mathbf{v}_k^T\mathbf{v}_k)$ before stacking.
Stacking $N$ flow measurements yields the system $\mathbf{b} = A\boldsymbol{\omega}$ where $\mathbf{b} \in \mathbb{R}^{3N}$ contains the projected velocities and $A \in \mathbb{R}^{3N \times 3}$ contains the projected observation models.
Least squares then gives the coarse spin vector that seeds \scmax: brute-force magnitude search (Suppl.~Fig.~7), then axis-and-magnitude refinement.

\subsubsection{s-CMax spin estimation refinement}
\label{sec:method_scmax_spin_estimation}

We adopt the contrast maximization framework \cite{gallego_unifying_2018} to estimate the ball's angular velocity $\boldsymbol{\omega} \in \mathbb{R}^3$ from a set of $N_e$ events $\{e_k\}^{N_e}_{k=1}$.
However, unlike \cite{nakabayashi_event-based_2024}, we use a spherical formulation of this framework specifically for angular motion estimation which we call \scmax to differentiate it from the formulation used for angular \textit{and} linear motion estimation.
Each event is represented by a tuple $e_k = (\mathbf{u_k}, t_k, p_k)$, where $\mathbf{u}_k = \left[u_k, v_k\right]^\intercal \in \mathbb{N}^2_0$ is the 2D pixel location, $t_k \in \mathbb{R}$ the timestamp, and $p_k \in \{0, 1\}$ the polarity (either OFF or ON, respectively).

We use the ball detection result obtained at time $t_\text{ref}$ (Sec.~\ref{sec:ball_detection}) to discard background events and lift the 2D pixel coordinates to 3D points $\mathbf{x}_k = \left[x_k, y_k, z_k\right]^\intercal$ on the unit sphere.
Assuming an orthographic projection, this lifting is computed as
\begin{equation}
\begin{aligned}
    x_k &= (u_k - c_x) / r, \quad
    y_k = (v_k - c_y) / r, \quad
    z_k = -\sqrt{1 - (x_k^2 + y_k^2)}
\end{aligned}  
\label{eq:orthographic}
\end{equation}
where $(c_x, c_y)$ is the detected ball center and $r$ is its radius in pixels.

Each 3D event point $\mathbf{x}_k$ is warped to $\mathbf{x}'_k = f_\text{warp}(\mathbf{x}_k, t_k, \boldsymbol{\omega})= R(-\boldsymbol{\omega} \cdot \Delta t_k) \, \mathbf{x}_k$ by rotating it according to the candidate spin,
where $\Delta t_k = t_k - t_\text{ref}$ and $R(\cdot) \in \mathrm{SO}(3)$ is the rotation matrix constructed from the axis-angle representation.

We convert the warped 3D points to unit spherical coordinates $\mathbf{s}'_k = [\theta'_k, \phi'_k]$ where $\theta'_k = \arcsin(z'_k)$ and $\phi'_k = \mathrm{arctan2}(y'_k, x'_k)$ denote the polar and azimuth angles, respectively.
The \iwe $I$ is obtained by accumulating events into a 2D histogram over discrete spherical coordinates
\begin{equation}
\label{eq:image_warped_events}
I[\mathbf{s}] = \sum_{k=1}^{N_e} \delta(\mathbf{s} - \mathbf{s}'_k)    
\end{equation}
where $\delta(\mathbf{x})$ is the Kronecker delta (1 if $|\mathbf{x}| = 0$, else 0).
The spin estimate is found by maximizing the variance of the \iwe
\begin{equation}
\label{eq:variance}
    \boldsymbol{\omega}^* = \argmax_{\boldsymbol{\omega}} \mathrm{Var}(I(\boldsymbol{\omega})).
\end{equation}

We solve this optimization problem in two steps: 1) we brute force search the spin magnitude and then 2) further refine it using the Nelder-Mead simplex algorithm \cite{gao_implementing_2012} as we find it yields better solutions than gradient-based methods in fewer iterations.
We use bilinear interpolation when accumulating events into the \iwe and smooth it with a $5\times5$ Gaussian kernel and unit standard deviation to facilitate convergence.
It is worth highlighting that the use of a spherical instead of planar projection of the \iwe~\cite{nakabayashi_event-based_2024} enables us to warp events to their true location on the ball without ambiguities.
We found that a non-unique planar projection of the \iwe leads to alignment of events from disparate locations on the ball surface, as well as the tendency for events to concentrate on the ball edge, which leads to high-variance configurations that do not reflect the underlying motion and thus bias the spin estimates.

\subsection{Online spin estimation}
\label{sec:method_online_spin_estimation}

\subsubsection{Online \cnn for spin estimation}
\label{sec:method_cnn_spin_estimation}

While \scmax yields accurate spin estimates, its iterative optimization precludes real-time operation.
Optical flow is faster but degrades under limited texture visibility and exhibits event-rate-dependent latency, and neither method provides an interpretable uncertainty estimate.
To address these limitations, we train a custom \cnn based on ResNet-18~\cite{he_deep_2016} to predict a 3D spin vector and a 3D variance vector that captures prediction uncertainty \cite{kendall_what_2017}.
We regress the angular velocity components, expressed in the virtual camera frame.
During training, we normalize the target by the maximum spin magnitude observed in the dataset and train the network to predict  $\hat{\tilde{\boldsymbol{\omega}}}$.

For balls with a spatially concentrated texture, like a logo (e.g., table tennis or golf), the texture may not be observable by the \gcs within the input time window.
Through the heteroscedastic loss, the network learns to represent uncertainty arising from limited texture visibility, translational motion, occlusions, and sensor noise without requiring explicit uncertainty labels, which would be cumbersome and ambiguous to produce.

The \cnn takes as input a normalized polarity-separated event time surface accumulated over \SI{15}{\milli\second}, which is chosen such that, even at low spin rates, a sufficient fraction of a revolution is observed.
This representation is masked using the predicted bounding box from the ball detection network (Fig.~\ref{fig:method_overview}).
In addition to the spin estimate $\hat{\tilde{\boldsymbol{\omega}}}$, the network predicts a per-component log-variance vector $\mathbf{s}=(s_x,s_y,s_z)$, where $s_i=\log \sigma_i^2$.
Predicting log-variances improves numerical stability and ensures strictly positive variances.
We train the model using a heteroscedastic Gaussian negative log-likelihood with diagonal covariance:
$$\mathcal{L}(\hat{\tilde{\boldsymbol{\omega}}},\mathbf{s})=\frac{1}{2}\sum_{i\in{x,y,z}}\left(\exp(-s_i)\left(\tilde{\omega}_i-\hat{\tilde{\omega}}_i\right)^2+s_i\right).
$$
This formulation encourages the network to increase predicted uncertainty when the input is ambiguous (e.g., due to occlusions or missing logo evidence), while remaining confident when the spin is well observed.
For training, we label a dataset of free-flight shots obtained from a ball thrower and real, elite-level table tennis games, using offline \scmax. 
We use a total of $3969$ event recordings, each corresponding to a single free-flight shot ($2233$ for training, $686$ for validation, and $1050$ for testing; roughly \SI{4}{\hour} of data in total), with magnitudes ranging from 0--9\,\si{\kilo\rpm} (see Suppl.~Tab.~3 for the distribution of the throws across magnitude ranges).
To improve robustness, we apply augmentations during training, including random noise injection, event dropping, temporal jitter, and spatial translation. 
We additionally apply random in-plane rotations to the event representation and rotate the ground-truth spin axis to improve axis coverage. 
Note that, although the spin vector is constant during free-flight, the spin axis expressed in the (virtual) camera frame changes over time due to the mirror motion of the \gcs.
Adapting the \cnn to a new ball requires sport-specific training but no manual labels: one records event data and runs offline \scmax to obtain pseudo-ground-truth.

\subsubsection{Online \scmax for batch spin refinement}
\label{sec:online_cmax_spin_estimation}

While the \cnn provides robust spin axis predictions that align well with the pseudo-ground-truth, we observe that its magnitude estimates tend to be conservative in high-spin scenes, likely due to the limited availability of training data in this regime (Suppl.~Tab.~3). 
To leverage the complementary strengths of learning-based and model-based approaches, we develop an online batch version of \scmax to asynchronously refine the spin magnitude, with the \cnn output as a reliable directional prior. 
Since we want to avoid time-consuming memory allocations in the decoder thread of the \evs, we convert the time surface representation back to events which we found to be fast and effective.
However, due to latency constraints, rather than iteratively optimizing the spin vector as in the offline version (Sec.~\ref{sec:method_offline_spin_estimation}), we opt to explore all reasonable magnitudes for a given axis in parallel on the GPU (see Suppl.~Sec.~A.6 for details).
For each batch, we evaluate the best GPU-searched magnitude and the original \cnn magnitude with the same \iwe-variance objective.
If the searched candidate yields a higher variance, we keep the \cnn axis and replace only the magnitude.

\section{Experiments}
\label{sec:experiments}

We evaluate real-world spin estimation accuracy using two primary metrics: 1) spin magnitude error, expressed as a percentage (\%) or in revolutions per minute (\si{\rpm}), and 2) spin axis error in degrees (\si{\degree}).
A fundamental challenge in evaluating spin estimation methods is the difficulty of obtaining reliable ground truth: attaching sensors or markers to a ball in free flight alters its dynamics, and frame-based vision references degrade with distance, motion blur, and temporal aliasing at high spin rates.
We therefore design three experiments that progressively trade controlled ground truth for realism: a static ball spinner with encoder-based ground truth (Sec.~\ref{sec:exp_spin_estimation_ball_spinner}), a ball thrower with vision-based pseudo-ground-truth labels (Sec.~\ref{sec:exp_spin_estimation_ball_thrower}), and elite-level table tennis rallies where ground truth is only known via the methods presented in this work (Sec.~\ref{sec:results_spin_estimation_free_flight}).

\subsection{Experimental setup}
\label{sec:exp_setup}

We run the \gcs on a dedicated computer with a multi-core CPU and two GPUs: the frame generation, spin filter, and ball state estimator run asynchronously on the CPU, while the ball detection, \cnn, and \scmax modules use separate GPUs to decrease latency and increase throughput.
The online components are written in C++ and CUDA, using TensorRT for neural network acceleration (see Suppl.~Sec.~A.1 for details).
For the ball thrower (Sec.~\ref{sec:exp_spin_estimation_ball_thrower}) and table tennis match (Sec.~\ref{sec:results_spin_estimation_free_flight}) experiments, we use a \bpms that covers a $\SI{14}{\meter} \times \SI{7}{\meter} \times \SI{5}{\meter}$ volume with about \SI{3.0}{\milli\meter} accuracy, streaming ball positions via ROS2 at \SI{200}{\hertz} with \SI{7.5}{\milli\second} latency.
All hardware is off-the-shelf (Prophesee EVK4, Thorlabs GVS012/M galvanometers, and an Optotune EL-10-30 lens) except for two 3D-printed mounts; we will publicly release our event datasets, pseudo-ground-truth labels, pseudocode, and a bill of materials with CAD files.

\subsection{Offline spin estimation experiments: ball spinner}
\label{sec:exp_spin_estimation_ball_spinner}

\ifshowfigures
\begin{figure}[!t]
    \centering
    \includegraphics[width=0.8\linewidth]{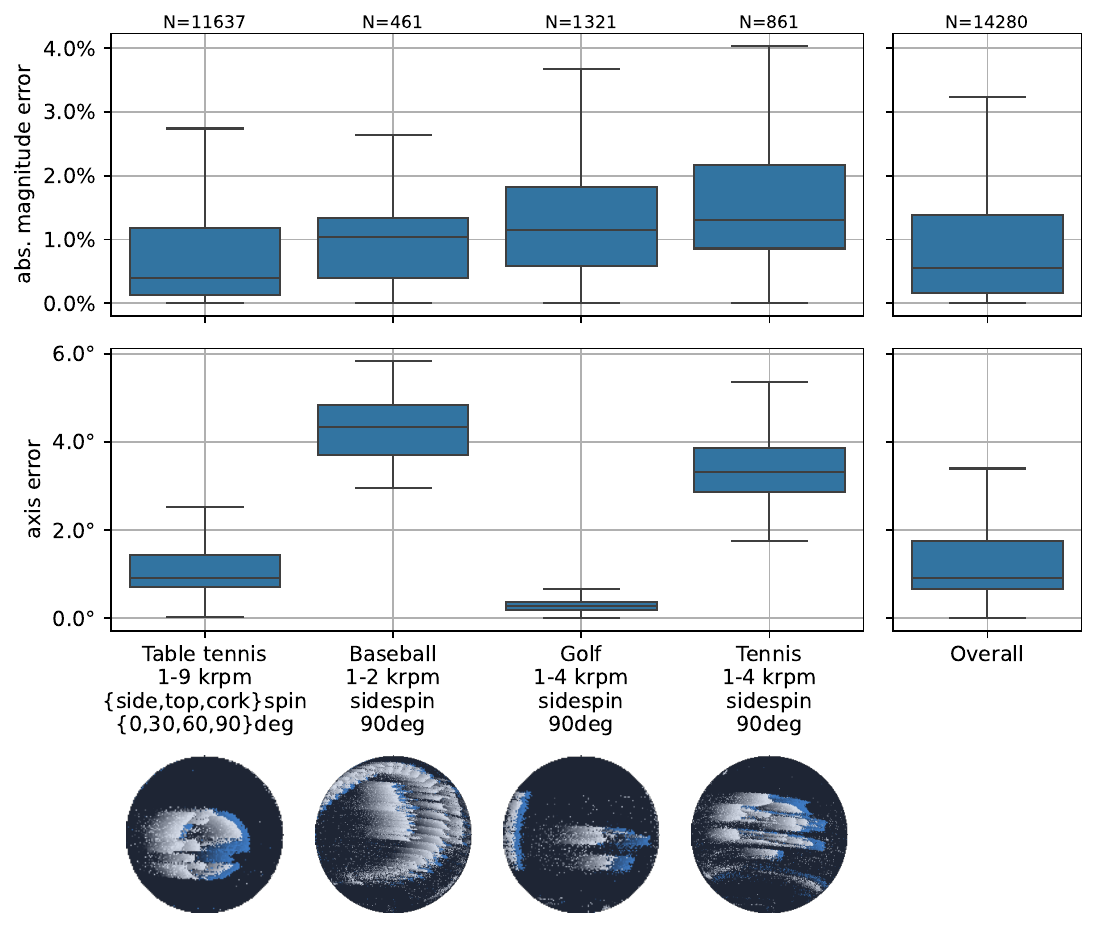}
    \caption{
    The effect of different ball types on the \scmax magnitude and axis error.
    The bottom row shows a time surface representation of the different balls when their pattern is visible, spinning at \SI{1}{\kilo\rpm}.
    }
    \label{fig:four_balls}
\end{figure}
\fi
We use motorized ball spinners to collect real-world spin estimation data with ground-truth axis and magnitude labels.
The spin magnitude is obtained from the motor encoders, whereas the spin axis is obtained by detecting an AprilTag \cite{wang_apriltag_2016} that is aligned with the spin axis of the motor (see Suppl.~Sec.~B.4 for details).

We recorded $10$ raw event files in a sidespin configuration across three ball types: $4$ files each for the golf and tennis ball from 1--4\,\si{\kilo\rpm}, and $2$ files for the baseball from 1--2\,\si{\kilo\rpm}, all in \SI{1}{\kilo\rpm} increments.
We additionally recorded $108$ table tennis-specific raw event files covering three spin types (topspin, sidespin, and corkspin), four logo orientations (0--90\,\si{\degree} in \SI{30}{\degree} increments), and nine spin magnitudes from 1--9\,\si{\kilo\rpm} in \SI{1}{\kilo\rpm} increments.

Unless otherwise noted, we compute spin estimates with an event accumulation time of \SI{10}{\milli\second}.
In order to exclude samples in which no events are generated by the pattern on the ball, e.g., when the pattern is rotated away from the observer, we set a $0.1$ minimum variance threshold.

\subsubsection{s-CMax spin estimation}
\label{sec:cmax_ball_types}

We conduct a ball spinner experiment to show the generalization of \scmax to different visual appearances (Fig.~\ref{fig:four_balls}).
We obtain similar magnitude and axis errors for all the ball types we tested in our experiments.
Some variations, e.g., the slightly higher axis error of around \SI{4}{\degree} for the baseball and tennis ball, can be attributed to setup inaccuracies: the heavier balls must be spun at lower magnitudes for safety and their AprilTag-based axis alignment on the spinner is less precise (see Suppl.~Sec.~B.4 for details).

\subsubsection{Comparison to other spin estimation methods}
\label{sec:comparison}

We compare our spin estimation methods against a flow-based \cite{gossard_table_2024} and a contrast-maximization-based baseline \cite{nakabayashi_event-based_2024}.
To ensure a fair comparison between methods, we make some modifications to the baselines (see Suppl.~Sec.~B.7 for details).
The flow-based method \cite{gossard_table_2024} depends on estimating the spin magnitude via peak finding in the event rate to set event accumulation time and minimum/maximum optic flow speed constraints.
The CMax-based method \cite{nakabayashi_event-based_2024} depends on random initialization of the spin vector, which we replace by the deterministic result from our optic flow method (Sec.~\ref{sec:method_flow_spin_estimation}).
We further adapt its parameters to match \scmax where they yield better results on our dataset: rotation-only estimation, a larger $5 \times 5$ Gaussian kernel for blurring the \iwe, and the same convergence criterion.

The results show clear improvements from each of our contributions (Tab.~\ref{tab:method_comparison_new_details}).
Our flow-based method, which serves as the initialization for the CMax-based methods, already halves the magnitude error and reduces the axis error by $4\times$ compared to Gossard~\etal~\cite{gossard_table_2024}, providing a stronger starting point for the subsequent optimization.
Building on this initialization, \scmax achieves the best accuracy across both metrics ($1.2\%$ magnitude and \SI{1.5}{\degree} axis error) while requiring only about one ninth of the runtime of CMax~\cite{nakabayashi_event-based_2024}, striking a favorable balance between accuracy and computational cost.

\ifshowtables
\begin{table}[ht]
\centering
\caption{Comparison of spin estimation methods on the spinner dataset. 
Magnitude and axis errors are reported as mean $\pm$ standard deviation.}
\label{tab:method_comparison_new_details}
\begin{tabular}{l
  S[table-format=2.1(2.1)]
  S[table-format=2.1(2.1)]
  S[table-format=3.1(3.1)]}
\toprule
\textbf{Method} & {\textbf{Mag. error} [\si{\percent}] $\downarrow$} & {\textbf{Axis error} [\si{\degree}] $\downarrow$ } & {\textbf{Runtime} [\si{\milli\second}] $\downarrow$ } \\
\midrule
Flow~\cite{gossard_table_2024}                & 35.2 \pm 19.5     & 37.1 \pm 24.2     & 7.7 \pm 14.1 \\
Flow (ours)                                   & 17.3 \pm 22.6     & 9.0 \pm 17.6      & \boldsymbol{$7.3 \pm 13.3$} \\
\midrule
CMax~\cite{nakabayashi_event-based_2024}      & 8.2 \pm 14.3      & 5.0 \pm 10.4      & 317.4\pm 425.3 \\
\scmax (ours)                                 & \boldsymbol{$1.2 \pm 2.3$}       & \boldsymbol{$1.5 \pm 2.2$}       & 36.7 \pm 19.5 \\
\bottomrule
\end{tabular}
\end{table}
\fi

\subsection{Offline spin estimation experiments: ball thrower}
\label{sec:exp_spin_estimation_ball_thrower}

\ifshowtables
\begin{table}[ht]
\centering
\caption{Comparison of spin estimation methods on the dotted table tennis ball on the ball thrower dataset. 
Magnitude and axis errors are reported as mean $\pm$ standard deviation. 
A large axis error (\textcolor{red}{red numbers}) indicates complete failure, even when the magnitude error is low.
Pseudo-ground-truth labels were obtained using SpinDOE~\cite{gossard_spindoe_2023}.}
\label{tab:wide_angle_comparison}
\begin{tabular}{ll
  S[table-format=2.1(2.1)]
  S[table-format=2.1(2.1)]}
\toprule
\textbf{Sensor setup} & \textbf{Method} & {\textbf{Mag. error} [\si{\percent}] $\downarrow$} & {\textbf{Axis error} [\si{\degree}] $\downarrow$ } \\
\midrule
Static \evs & Flow~\cite{gossard_table_2024} & 18.8 \pm 12.9 & \textcolor{red}{$88.7 \pm 8.1$} \\
Static \evs & Flow (ours)       & 14.1 \pm 8.7     & \textcolor{red}{$59.9 \pm 5.0$} \\
\midrule
Static \evs & CMax~\cite{nakabayashi_event-based_2024} & n/a    & n/a \\
Static \evs & \scmax (ours)        & n/a     & n/a \\
\midrule
\gcs (ours) & Flow~\cite{gossard_table_2024}       &  14.7 \pm 12.6 & \textcolor{red}{$75.6\pm 18.5$ }\\
\gcs (ours)& Flow (ours)                       & 18.0 \pm 5.4       & 11.8 \pm 29.4 \\
\midrule
\gcs (ours) & CMax~\cite{nakabayashi_event-based_2024} & 10.0 \pm 20.0  & 12.3 \pm 29.3 \\
\gcs (ours) & \scmax (ours)                       & \boldsymbol{$2.1 \pm 2.5$}       & \boldsymbol{$5.4 \pm 3.7$} \\
\bottomrule
\end{tabular}
\end{table}
\fi

We place a \gcs next to a static \evs with a wide-angle lens---a common configuration in related works \cite{nakabayashi_event-based_2024,gossard_table_2024}---both \SI{2}{\meter} from the ball thrower, so that both observe the same $30$ yellow dotted shots from the same perspective (see Suppl.~Sec.~B).
This isolates the two key benefits of the \gcs over a static wide-angle camera: a magnified, higher-resolution view of the ball surface, and compensation of linear motion via active tracking.

The results show that it is possible to obtain reliable spin estimates with the \gcs in this setting, whereas the static \evs cannot provide satisfactory results (Tab.~\ref{tab:wide_angle_comparison}).
There are two main reasons for this discrepancy: 1) linear motion that triggers events and 2) limited resolution on the ball surface.
For the flow-based methods, the optic flow estimates follow the ball motion rather than the spin and closely match the image velocity found by computing finite differences from subsequent ball detections, indicating that only the linear ball motion can be extracted despite considerable spin.
For the CMax-based methods, we did not obtain results beyond the initial guess provided by the flow method due to an insufficient number of events on the ball, suggesting that the resolution is too limited for these methods.
The seemingly reasonable errors on the flow methods for the static \evs are coincidental: the ball thrower couples speed and spin so translational motion produces magnitudes numerically similar to that of the true spin – but with the axis wrong, the magnitude is meaningless.
This is precisely the translation-rotation conflation the \gcs resolves.

\subsection{Online spin estimation experiments: real table tennis games}
\label{sec:results_spin_estimation_free_flight}

We evaluate the \cnn on the held-out test split (1050 recordings) to assess spin prediction accuracy and the quality of the uncertainty estimates.
Fig.~\ref{fig:spin_estimation_cnn_test_results} analyzes error as a function of predicted uncertainty reported by spin magnitude range. 

Across all ranges, the error decreases as we restrict evaluation to more confident predictions, indicating that the predicted uncertainty is informative for identifying reliable spin estimates. 
Both the median error and the spread increase for higher spin magnitudes, consistent with reduced training coverage in these regimes. 
In total, the mean error on the 50\% uncertainty percentile across all ranges is $8.8 \pm 12.4 \%$ for the magnitude and $6.4 \pm 6.7$ degrees for the axis.

\ifshowfigures
\begin{figure}[!t]
    \centering
    \includegraphics[width=\linewidth]{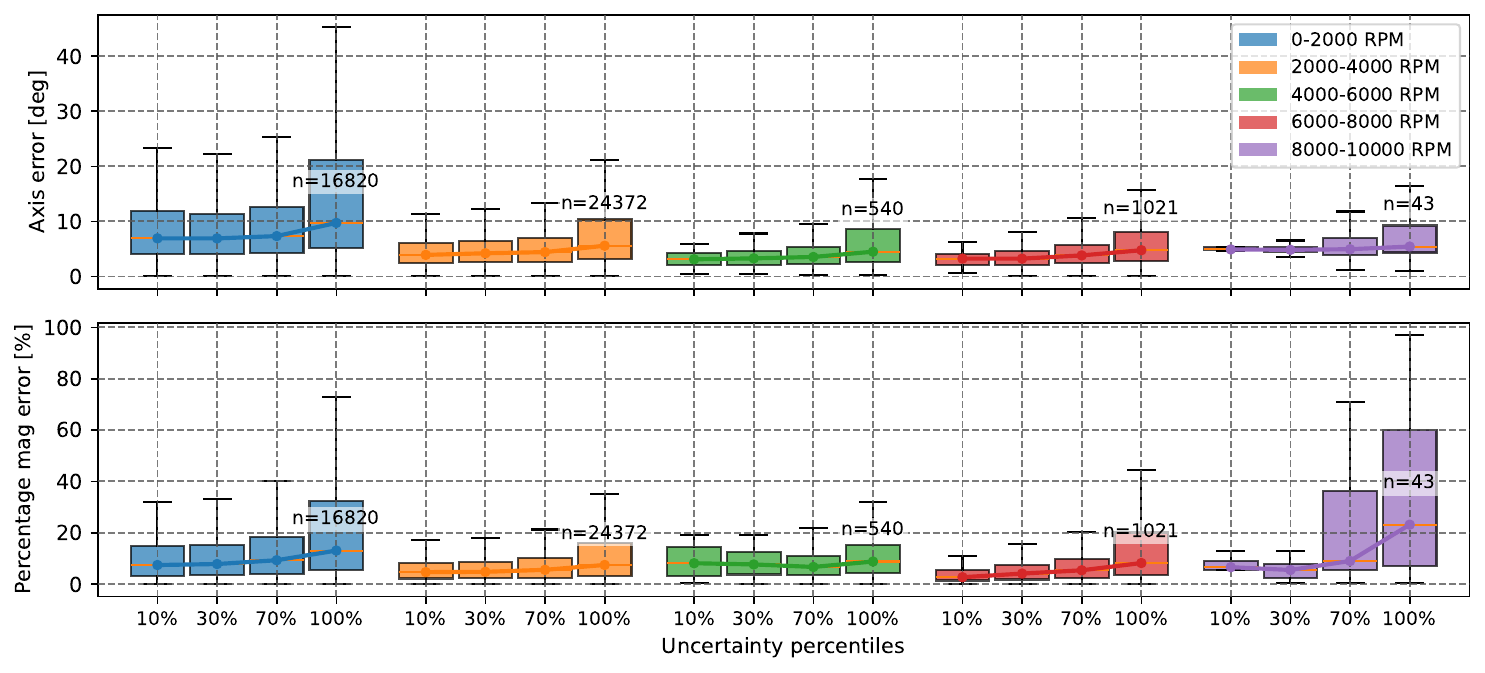}
   \caption{
       CNN magnitude and axis errors on the held-out test set for different uncertainty percentiles and magnitude ranges.}
    \label{fig:spin_estimation_cnn_test_results}
\end{figure}
\fi

The dominant failure mode is logo invisibility: across $2642$ free-flight segments from $7$ elite table tennis matches, \SI{2.2}{\percent} had the logo hidden for more than half of the duration, and the predicted uncertainty enables graceful degradation through multi-view fusion (Suppl.~Sec.~A.7).
A qualitative visualization (Suppl.~Sec.~B) confirms that clearly visible logo traces yield accurate, low-uncertainty predictions, whereas limited logo evidence increases both error and predicted uncertainty.

\section{Conclusion}
\label{sec:conclusion}

We presented an event-based active vision system for real-time ball tracking and spin estimation, combining an event camera with a focus-tunable telephoto lens and fast galvanometer mirrors to obtain a magnified, motion-compensated view of unmodified balls in flight.
The hybrid tracking framework fuses 3D ball localization with 2D event-based detection for smooth, robust active tracking at professional ball-sport speeds.
The offline spin estimation method, based on a spherical contrast maximization formulation, achieves state-of-the-art accuracy on a variety of officially approved balls, outperforming existing baselines.
For real-time operation, we introduced an uncertainty-aware CNN trained on pseudo-ground-truth labels from the offline method, combined with a GPU-accelerated batch contrast maximization refinement step and a multi-view fusion strategy, achieving state-of-the-art latency and throughput during professional table tennis matches.
We believe the system's ability to track and resolve fine surface details at high temporal resolution opens promising avenues for sports broadcasting, high-speed object inspection, and competitive robotic play.

\section*{Acknowledgments}
\label{acknowledgements}
We thank 
Dario Brescianini for initiating the project,
Markus Kamm for the custom lens optics,
Robin Frauenfelder and Carter Fang for the initial ball tracking and detector prototype,
Alexander Sigrist, Stefan Heusser and Nobuhiko Mukai for building the camera synchronization hardware, ball spinners, and ball throwers,
Christian Conti Fujiwara for the ball aerodynamic model and the contact physics models,
Christian Conti Fujiwara and Hamdi Sahloul for the SpinDOE implementation,
Valentin Monferrato and Adriano Patane for data labeling and cleaning,
and Aleksandar Stoimenov for debugging and fixing technical issues.

\bibliographystyle{splncs04}
\bibliography{main}



\end{document}